\documentclass{article}
\usepackage{spconf,amsmath,graphicx}
\usepackage{bm}
\usepackage{color}
\usepackage{url}
\usepackage{fancyvrb}
\usepackage[subrefformat=parens]{subcaption}

\title{STREAMING JOINT SPEECH RECOGNITION AND DISFLUENCY DETECTION}

\name{
\begin{tabular}{c}
Hayato Futami$^1$, Emiru Tsunoo$^1$, Kentaro Shibata$^1$, Yosuke Kashiwagi$^1$, Takao Okuda$^1$, \\ 
Siddhant Arora$^2$, Shinji Watanabe$^2$
\end{tabular}
}
\address{$^1$Sony Group Corporation, Japan $^2$Carnegie Mellon University, USA}

\begin{document}
\ninept
\maketitle

\begin{abstract}
Disfluency detection has mainly been solved in a pipeline approach, as post-processing of speech recognition.
In this study, we propose Transformer-based encoder-decoder models that jointly solve speech recognition and disfluency detection, which work in a streaming manner.
Compared to pipeline approaches, the joint models can leverage acoustic information that makes disfluency detection robust to recognition errors and provide non-verbal clues.
Moreover, joint modeling results in low-latency and lightweight inference.
We investigate two joint model variants for streaming disfluency detection: a transcript-enriched model and a multi-task model.
The transcript-enriched model is trained on text with special tags indicating the starting and ending points of the disfluent part.
However, it has problems with latency and standard language model adaptation, which arise from the additional disfluency tags.
We propose a multi-task model to solve such problems, which has two output layers at the Transformer decoder; one for speech recognition and the other for disfluency detection.
It is modeled to be conditioned on the currently recognized token with an additional token-dependency mechanism.
We show that the proposed joint models outperformed a BERT-based pipeline approach in both accuracy and latency, on both the Switchboard and the corpus of spontaneous Japanese.
\end{abstract}
\vspace{-10pt}
\section{Introduction}
\vspace{-5pt}
Disfluencies are interruptions in the normal flow of spontaneous speech that do not appear in written text.
\begin{Verbatim}[fontsize=\small]
Flights [from Boston + {uh I mean} to Denver]
\end{Verbatim}
The above example shows a disfluency structure introduced in \cite{Shriberg94-PTSD}.
A reparandum ``from Boston'' is a part to be repaired.
An optional interregnum ``uh I mean'' (such as filled pauses and discourse markers) follows after an interruption point marked as ``+.''
Then, a repair part ``to Denver'' follows.
Both reparanda and interregna are regarded as disfluencies in this study.

Automatic speech recognition (ASR) systems usually transcribe such disfluencies as they are, which degrades the human readability of transcripts.
They can also adversely affect downstream natural language processing (NLP) tasks, such as machine translation \cite{Salesky18-TFT}, question answering \cite{Gupta21-disflQA}, and intent classification \cite{Dao22-FDD}.
To address these issues, many studies have been conducted to detect and remove disfluencies.
Most existing studies have adopted a pipeline approach, where an ASR model converts speech into text, and a disfluency detection model then tags tokens in this text.
As in many NLP tasks, BERT \cite{Devlin19-BERT} has become a popular choice for disfluency detection \cite{Lou20-IDD, Rocholl21-DDUD}.
Such pipeline approaches usually work offline, assuming complete input text.
As ASR is often conducted in online streaming processing, a streaming disfluency detection method is desirable, which works immediately after streaming ASR outputs a partial result.
Recently, some studies have applied BERT to partial text incrementally \cite{Rohanian21-BBW, Chen22-TBW}.

In this study, we consider a streaming joint model for ASR and disfluency detection.
Among some options for streaming ASR modeling, we adopt a Transformer-based model because it is compatible with a token-level NLP task.
The joint model utilizes acoustic features, which makes the model robust against ASR errors.
In addition, non-verbal clues for disfluency detection can be used, such as interruption points \cite{Zayats19-GAU}.
Finally, the model also has the advantages of low latency and a small memory footprint.

To jointly solve ASR and disfluency detection, a transcript-enriched model can be considered.
The transcript-enriched model is trained on enriched transcripts that have disfluency annotations, where special symbols for the starting and ending points of the disfluent part are added.
This style of modeling is popular for solving some spoken language understanding (SLU) tasks \cite{Omachi21-EASRJP, Arora22-ESPSLU}, and similar non-streaming variants have already been explored in non-streaming disfluency detection \cite{Omachi22-NAE}.
However, the latency will increase because of the increased sequence length with special symbols.
In addition, conventional language models (LMs) cannot be straightforwardly applied, because we have to handle these special symbols included in their outputs.
To overcome these shortcomings, we propose a multi-task model that has an additional output layer to streaming Transformer ASR to predict whether the next token is disfluent or not.
We use disfluency detection history along with transcript history as input.
We also add a novel token-dependency mechanism, to model dependencies between token and disfluency predictions.
We experimentally confirmed the effectiveness of these joint models on both Switchboard and the corpus of spontaneous Japanese (CSJ) that have disfluency annotations.

\vspace{-10pt}
\section{Related work}
\vspace{-5pt}

\vspace{-3pt}
\subsection{Streaming end-to-end ASR}
\vspace{-3pt}
\label{sec:streaming-e2e-asr}
In streaming end-to-end ASR, both its encoder and decoder should operate online.
On the encoder side, particularly for a Transformer encoder, blockwise processing has been shown to be effective, where the encoder processes fixed-length speech blocks at certain time intervals.
Some studies have proposed contextual block processing \cite{Tsunoo19-TACBP, Shi21-EMF} to inherit the global context information.

On the decoder side, there are two major streams: frame-synchronous and label-synchronous models.
Frame-synchronous models, such as RNN-T \cite{He19-SESR}, provide frame-level predictions.
Label-synchronous models, such as attention-based decoder \cite{Tsunoo21-STABS, Chiu18-MCA}, provide token-level predictions.
As disfluency detection is a token-level tagging task, a label-synchronous model with a Transformer decoder will be natural to solve it.
Label-synchronous models need to detect the decoding end of the current block, and several ways for that have been proposed \cite{Tsunoo21-STABS, Chiu18-MCA, Niko19-TA}.
In this study, we adopt \cite{Tsunoo21-STABS} because it performs comparably without any additional training.

\vspace{-3pt}
\subsection{Disfluency detection}
\vspace{-3pt}
\label{sec:related-disfluency}
Disfluency detection has been mainly conducted offline and solved as post-processing of ASR, which we refer to as a ``pipeline'' method.
Prior to Transformer, it has been solved as a sequence labeling problem using RNNs (LSTMs) and their variants \cite{Wang17-TDD, Lou18-DDACNN, Bach19-NBLSTM}.
Since the rise of Transformer, it has started to be solved as a translation task using encoder-decoder models \cite{Wang18-Semi, Dong19-AT}.
As in many NLP tasks, the most successful approach now is to fine-tune pre-trained models such as BERT \cite{Lou20-IDD, Rocholl21-DDUD, Bach19-NBLSTM}.

\begin{figure}[t]
  \begin{minipage}[b]{0.48\linewidth}
    \centering
    \includegraphics[keepaspectratio, scale=0.52]{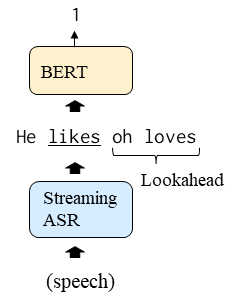}
    \subcaption{Pipeline}
  \end{minipage}
  \begin{minipage}[b]{0.48\linewidth}
    \centering
    \includegraphics[keepaspectratio, scale=0.52]{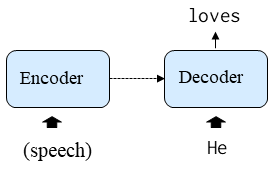}
    \subcaption{End-to-end}
  \end{minipage}
  \caption{(a) Pipeline method with BERT for streaming disfluency detection and (b) end-to-end disfluency removal models. The example text is ``He likes oh loves her'' where ``oh'' and ``loves'' are disfluencies (labeled as $1$).}
  \label{fig:related}
  \vspace{-10pt}
\end{figure}

Recently, some studies have addressed streaming, or incremental, disfluency detection \cite{Rohanian21-BBW, Chen22-TBW}, where detection is conducted on incomplete text, as early as possible after being output from streaming ASR.
To solve the input mismatch between training and inference, BERT has been trained not only on complete text but also on prefixes of them.
In addition, a lookahead window, whose size is fixed or adaptive \cite{Chen22-TBW}, plays an important role in accurate detection.
The streaming disfluency detection with BERT is shown in Fig. \ref{fig:related} (a).

Disfluency detection has been jointly solved with dependency parsing \cite{Lou20-IDD} or other NLP tasks \cite{Lee21-ASL}.
It has been solved with ASR \cite{Omachi22-NAE, Inaguma18-EAJSSD}, where CTC or mask CTC was trained on enriched transcripts with special tags.
However, streaming processing has not been considered in these studies.

Besides disfluency detection, some studies have investigated end-to-end disfluency removal, which directly predicts transcripts without disfluencies from speech \cite{Jamshid20-ESRDR, Horii22-EESSR}, as shown in Fig. \ref{fig:related} (b).
From a practical point of view, such models are occasionally inapplicable, because they cannot provide transcripts for the disfluent part.
Disfluencies carry some speakers' emotions, such as a lack of confidence, which are useful for certain applications.
Moreover, when an excessive removal occurs, the corresponding transcript will not be available by any means, while available by controlling a threshold in detection methods.

\vspace{-10pt}
\section{Joint modeling of Streaming Speech recognition and Disfluency detection}
\vspace{-5pt}

\begin{figure}[t]
  \begin{minipage}[b]{0.48\linewidth}
    \centering
    \includegraphics[keepaspectratio, scale=0.55]{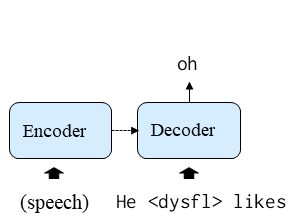}
    \subcaption{Transcript-enriched}
  \end{minipage}
  \begin{minipage}[b]{0.48\linewidth}
    \centering
    \includegraphics[keepaspectratio, scale=0.55]{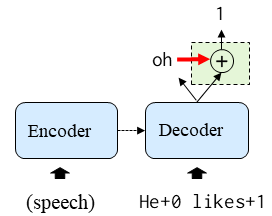}
    \subcaption{Multi-task}
  \end{minipage}
  \caption{Two variants of joint ASR and disfluency detection models. (a) Transcript-enriched model and (b) multi-task model. Green box indicates a token-dependency mechanism.}
  \label{fig:joint}
  \vspace{-10pt}
\end{figure}

In this study, we solve streaming disfluency detection by joint modeling with ASR.
Compared to pipeline methods, a streaming joint model can process both tasks in a single model, leading to latency and model size reduction.
It can also directly access speech information for disfluency detection.
Speech helps the model deal with recognized text with ASR errors.
In addition, speech carries extra information that is helpful for detecting disfluencies, which cannot be represented in text form.
Specific pauses, word durations, and intonation patterns are observed before or after the interruption point, as speakers realize that they have misspoken \cite{Zayats19-GAU, Shriberg99-PC}.
Furthermore, assuming blockwise streaming processing, the model can look at some future speech context before corresponding tokens are decoded.
We implement the joint model based on streaming Transformer ASR \cite{Tsunoo21-STABS} and investigate two variants of joint modeling as in Fig. \ref{fig:joint}.

\vspace{-3pt}
\subsection{Transcript-enriched model}
\vspace{-3pt}
A transcript-enriched model predicts text with disfluency tags from speech, as shown in Fig. \ref{fig:joint} (a).
We introduce special symbols for the starting and the ending points of the disfluent part marked as \url{<dysfl>} and \url{</dysfl>}.
Although other tagging strategies are possible \cite{Omachi22-NAE, Inaguma18-EAJSSD}, we adopt such a strategy to minimize the increase in sequence length.
Let $\bm{X}$ and $\bm{y} = (y_1, ..., y_{L})$ denote acoustic features and a corresponding token sequence of length $L$, respectively.
$y_i$ comes from the vocabulary $\mathcal{V}$, i.e., $y_i \in \mathcal{V}$.
Let $\bm{y}^{+} = (y^{+}_1, ..., y^{+}_{L'})$ denote an enriched token sequence of length $L'$ ($\geq L$), where $y^{+}_i \in \mathcal{V'} = \mathcal{V} \cap$ (\url{<dysfl>}, \url{</dysfl>}).
The training objective is formulated as follows:
\begin{align}
\mathcal{L} = - \sum_i \log p(y^{+}_i | \bm{X}, \bm{y}^{+}_{<i}),
\end{align}
where $\bm{y}^{+}_{<i}$ represents $(y^{+}_1, ..., y^{+}_{i-1})$.

\vspace{-3pt}
\subsection{Multi-task model}
\vspace{-3pt}
\label{sec:multi-task}
In the transcript-enriched model, disfluency tags are incorporated into token sequences $\bm{y}$ to make $\bm{y}^{+}$.
These tags increase the sequence length from $L$ to $L'$, which has a negative effect on the latency.
In addition, LMs are usually trained on the untagged text and do not work well with tagged outputs from the transcript-enriched model due to the vocabulary mismatch between $\mathcal{V}$ and $\mathcal{V'}$.
Although LMs can be trained on tagged text, such data are obtained from disfluency-annotated spoken text, which is often limited compared to untagged text.

To avoid using disfluency tags, we propose a multi-task model, which predicts the next token $y_i$ first, and then determines whether this token is disfluent or not, as shown in Fig. \ref{fig:joint} (b).
The decoder has two output layers for each role.
Let $\bm{d} = (d_1, ..., d_L)$ denote a disfluency sequence ($d_i = 1$ for disfluent tokens and $d_i = 0$ for fluent tokens) that corresponds to $\bm{y}$.
The joint training objective is formulated as the following factorized form:
\begin{align}
\label{eq:multi-task-loss}
\mathcal{L} &= - \sum_i \log p(y_i, d_i | \bm{X}, \bm{y}_{<i}, \bm{d}_{<i}) \nonumber \\
= & - \!\! \sum_i \log p(y_i | \bm{X}, \bm{y}_{<i}, \bm{d}_{<i}) \!
- \! \sum_i \log p(d_i | \bm{X}, \bm{y}_{\textcolor{red}{\leq i}}, \bm{d}_{<i}).
\end{align}
Note that $\bm{y}_{\leq i} = (\bm{y}_{< i}, y_i)$, which naturally incorporates the \textit{dependency} of predicting $d_i$ given the corresponding token $y_i$ through the above formulation based on joint probability distribution.
First, to model the \textit{dependency} between $d_i$ and previous disfluency predictions $\bm{d}_{<i}$, we input $\bm{d}_{<i}$ along with $\bm{y}_{<i}$.
To feed these input representations to the network, token and disfluency input embeddings are summed, similar to token and segment embeddings in BERT \cite{Devlin19-BERT}.

To capture the \textit{dependency} from $y_i$ to $d_i$ in Eq.~\eqref{eq:multi-task-loss}, we propose a token-dependency mechanism as an additional connection between the two output layers in the green box of Fig. \ref{fig:joint} (b).
Such a dependency is important because whether a token is disfluent or not is highly related to what the token is.
Note that our joint distribution formulation Eq.~\eqref{eq:multi-task-loss} can have another solution that makes token predictions $y_i$ depend on disfluency ones $d_i$, i.e., $p(y_i | \bm{X}, \bm{y}_{<i}, \bm{d}_{\textcolor{red}{\leq i}})$, which will be compared in Section \ref{sec:experimental-results}.
Let $y_i$, $E(y_i)$, and $\bm{s}_{i}$ denote the $i$-th (current) predicted token, its embedding, and the $i$-th final hidden state of the Transformer decoder, respectively.
For the $i$-th disfluency prediction, $E(y_i)$ and $\bm{s}_{i}$ are concatenated and linearly transformed as:\footnote{For simplicity, this formulation omits complicated notation of streaming processing. $y_i$ and $d_i$ are actually predicted before the entire $\bm{X}$ are obtained.}
\begin{align}
\label{eq:token-dependency}
p(d_i | \bm{X}, \bm{y}_{\textcolor{red}{\leq i}}, \bm{d}_{<i}) = {\rm softmax}(W[{\rm E}(y_i) ; \bm{s}_{i}] + b),
\end{align}
where $W$ and $b$ denotes the weight and bias of the linear layer, respectively.
During inference, we apply beam search with disfluency scores.
We expand hypotheses as the combinations of $\bm{y}_{\leq i}$ and $\bm{d}_{\leq i}$ based on the following score:
\begin{align}
\label{eq:disfluency-integrated-beam-search}
{\rm Score}(\bm{y}_{\leq i}, \bm{d}_{\leq i}) = - \sum_{j=1}^i \log p(y_j | \bm{X}, \bm{y}_{<j}, \bm{d}_{<j})  \nonumber \\
- \alpha \sum_{j=1}^i \log p(d_j | \bm{X}, \bm{y}_{\textcolor{red}{\leq j}}, \bm{d}_{<j}),
\end{align}
where $\alpha$ is a hyperparameter.
We found $\alpha = 1.0$ in the experiments.

\begingroup
\renewcommand{\arraystretch}{1.1}
\begin{table*}[t]
  \caption{Streaming ASR and disfluency detection on Switchboard without LM. ``WER'' denotes the word error rate including disfluencies, which is a standard ASR metric. ``DR-WER'' denotes disfluency removed WER. ``Aligned F1 / P / R'' denotes F1, precision, and recall score calculated by aligning ASR results to references. ``TL-Latency'' denotes the $50$th/$90$th percentile of token-level latency.}
  \vspace{-2mm}
  \label{tab:SWBD}
  \centering
  \begin{tabular}{lccccc} \hline
     & WER[\%] & DR-WER[\%] & Aligned F1 / P / R & TL-Latency (50/90) [ms] & Parameters \\ \hline
    ASR only \cite{Tsunoo21-STABS} & $18.1$ & $36.9$ & $-$ & $496/978$ & $50.6$M \\
    Joint (Multi-task) & $\bm{17.8}$ & $\bm{21.3}$ & $\bm{0.84}/0.89/0.79$ & $529/995$ & $58.4$M \\
    Joint (Transcript-enriched) & $17.9$ & $21.4$ & $\bm{0.84}/0.89/0.80$ & $575/1061$ & $50.6$M \\
    End-to-end removal \cite{Jamshid20-ESRDR} & $-$ & $21.7$ & $(0.78/0.78/0.78)$ & $\bm{482}/\bm{860}$ & $50.6$M \\ \hline
    Pipeline ASR + BERT (No Lookahead) & $18.1$ & $26.5$ & $0.69/0.80/0.60$ & $536/1045$ & $159$M \\
    Pipeline ASR + BERT (Lookahead=2) & $18.1$ & $22.6$ & $0.81/0.86/0.76$ & $641/1199$ & $159$M \\ \hline
 \end{tabular}
 \vspace{-10pt}
\end{table*}
\endgroup

\begingroup
\renewcommand{\arraystretch}{1.1}
\begin{table}[t]
  \caption{Ablation studies on components for multi-task model}
  \vspace{-2mm}
  \label{tab:ablation-study}
  \centering
  \begin{tabular}{lc} \hline
     & DR-WER[\%] \\ \hline
    Multi-task & $\bm{21.3}$ \\
    - token-dependency mechanism & $22.3$ \\
    \quad - input disfluency & $22.7$ \\
    Disfluency-dependency mechanism & $22.2$ \\ \hline
 \end{tabular}
 \vspace{-10pt}
\end{table}
\endgroup

\vspace{-10pt}
\section{Experimental evaluations}
\vspace{-5pt}

\subsection{Experimental conditions}
\vspace{-5pt}
\label{sec:experimental-conditions}
We conducted ASR and disfluency detection experiments on Switchboard 1 Release 2 (Switchboard) \cite{Godfrey92-SWBD} and the CSJ \cite{maekawa03-CSJ}.
Switchboard consists of 260 hours of approximately 2,400 English telephone conversations \cite{Godfrey92-SWBD}.
As a part of the Penn Treebank Release 3 dataset, 1,126 conversations were annotated for disfluencies \cite{Marcus99-TB3}.
Since the original transcript had some errors, Mississippi State University released corrected transcripts with utterance segmentation, which was necessary for ASR.
However, disfluency annotation did not exist in the corrected ``Msstate'' version.
Paired speech and text data with disfluency annotation were required to train joint ASR and disfluency detection models.
A previous study had labeled Msstate transcripts using a state-of-the-art disfluency detector \cite{Jamshid20-ESRDR}.
We believed that such labeling was too biased toward the detector.
We instead aligned disfluency-annotated original transcripts to segmented Msstate transcripts by backtracking the path of the smallest edit distance \footnote{We will release script: https://github.com/hayato-futami-s/joint-asr-dysfl}.
As a result, we obtained about 240k utterances (of 580k utterances in the Msstate version) after speed perturbation \cite{Ko15-AA}.
We used the same data splits as used in most previous studies: \url{sw[23]*} for training data, \url{sw4[5-9]*} for dev data, and \url{sw4[1-4]*} for test data.
Both reparanda and interregna were seen as disfluencies to be removed ($14\%$ of total tokens for training data), following \cite{Chen22-TBW}. 
CSJ consists of 520 hours of Japanese presentations \cite{maekawa03-CSJ}, which has disfluency annotations as \url{TagFiller}, \url{TagDisfluency},  and \url{TagDisfluency2} ($8.1\%$ of total tokens).

We built a streaming Transformer model for joint ASR and disfluency detection, which consisted of a 12-layer encoder and a 6-layer decoder using ESPnet \cite{Watanabe18-ESPnet}. 
The encoder was based on contextual block processing \cite{Tsunoo19-TACBP} with 1,600 ms block size and 640 ms shift.
The model was jointly trained and decoded with CTC of weight $0.3$.
During decoding, beam search of width $5$ was applied.
We also built an end-to-end disfluency removal model \cite{Jamshid20-ESRDR} in Fig. \ref{fig:related} (b), with the same architecture.
ASR was done in a wordpiece unit of size 30,522 on Switchboard and a character unit of size 3,261 on CSJ.

As a pipeline approach in Fig. \ref{fig:related} (a), we prepared BERT with ``Transformers'' library \footnote{https://huggingface.co/models}, \url{bert-base-uncased} for Switchboard and \url{cl-tohoku/bert-base-japanese-char-v2} for CSJ.
We fine-tuned them on text with disfluency labels.
For streaming disfluency detection, we fine-tuned BERT on all the prefixes \cite{Rocholl21-DDUD, Chen22-TBW}, as noted in Section \ref{sec:related-disfluency}.
For ASR on Switchboard, we use the same tokenizer as BERT to make a fair comparison.

We evaluated disfluency detection with ASR in terms of both accuracy and latency.
As an accuracy metric, we defined the disfluency removed word error rate (DR-WER).
We calculated WER on only fluent words, where hypotheses after removing disfluencies were compared with references without disfluencies.
While most previous studies in this field have reported F1 scores on the reference transcripts, we reported the performance on ASR results, where F1 scores could not be calculated straightforwardly.
We calculated F1 scores by aligning hypotheses with the references as ``aligned'' F1 score \footnote{Token inserted or deleted parts do not contribute to the F1 calculation.}.
However, it was affected by alignment mismatch, and it was difficult to say which was a disfluent part in the erroneously recognized text.
Therefore, we regards the DR-WER as the target metric for accuracy, while aligned F1 scores as an auxiliary metric.
As a latency metric, we calculated the token-level latency, i.e., the elapsed time from the acoustic boundary of a token to disfluency detection of the token, as noted in \cite{Inaguma21-AKD}.
The boundary was obtained from CTC segmentation \cite{Kurzinger20-CTCSeg}.
Latency calculation was conducted in teacher-forced decoding (beam width was $1$) \cite{Inaguma21-AKD}.

\vspace{-3pt}
\subsection{Experimental results}
\vspace{-3pt}
\label{sec:experimental-results}
First, we compared different models for streaming ASR and disfluency detection on Switchboard in Table \ref{tab:SWBD}.
The first column shows the original WER including disfluencies.
We found that both joint models did not hurt and even improved the WER, compared to the ASR-only model.
The end-to-end removal model output only fluent transcripts, so the WER including disfluencies could not be calculated.
Note that the WER became higher compared to the standard Switchboard setup with the Msstate transcripts, because of inaccurate original transcripts and smaller amounts of training data, as noted in Section \ref{sec:experimental-conditions}.
The second column shows the DR-WER.
For the ASR-only model, all the disfluencies remained and were counted as errors.
We found that both joint models outperformed the pipeline approachs, where looking ahead improved the DR-WER at the cost of the latency but still fell short of joint models.
Our joint models also outperformed the end-to-end removal model, with flexibility to choose fluent or disfluent hypotheses.
For the type of joint modeling, the multi-task model was comparable to the transcript-enriched model.
The third column shows aligned F1, precision (P), and recall (R) scores.
Similar trends to the DR-WER were observed.
Note that these values could not be directly compared to previous studies due to ASR error handling and utterance segmentation.
For the end-to-end model, ASR deletion errors and detection errors cannot be distinguished, which may not be a fair comparison to others.
The fourth column shows the $50$th and $90$th percentile of the token-level latency calculated on a Intel(R) Core(TM) i7-12700KF CPU.
The pipeline required one more step for BERT inference and waited for future tokens to achieve sufficient accuracy.
As a result, both joint models achieved lower latency than BERT with a $2$-token lookahead window.
Comparing the two joint models, the multi-task model was better at latency, because the output length of the transcript-enriched model was increased, as discussed in Section \ref{sec:multi-task}.
The end-to-end model had the lowest latency because its output length was decreased by removing disfluencies.
The final column shows the number of total parameters.
In the multi-task model, the token-dependency mechanism increased the model size, mainly because of the embedding layer for 30,522 vocabularies in Eq. (\ref{eq:token-dependency}).
However, both joint models were still much smaller than the BERT-based pipeline.

Next, we conducted ablation studies on the components for the proposed multi-task model in Table \ref{tab:ablation-study}.
We found that the token-dependency mechanism in Eq. (\ref{eq:token-dependency}) was necessary, as removing it degraded the DR-WER significantly, which shows the importance of the dependency of the current token $y_i$ to predict $d_i$.
We also found that it was important to input previous disfluency predictions besides token predictions to the model, as $\bm{d}_{<i}$ in Eq. (\ref{eq:multi-task-loss}).
Instead of the token-dependency mechanism, we added the disfluency-dependency mechanism as discussed in Section \ref{sec:multi-task}, whose performance was worse.

We evaluated the models with shallow fusion with LMs in Table \ref{tab:SWBD-LM}.
The training data for LMs were Switchboard transcripts and the external Fisher corpus \cite{Cieri04-Fisher}.
In addition to a standard RNNLM, we prepared a transcript-enriched RNNLM, where disfluency tags were added in the part of LM training data that has disfluency annotations.
Note that this requires an extra cost of training a special LM this is only applicable when in-domain disfluency-annotated data are available, as discussed in Section \ref{sec:multi-task}.
In the multi-task model, the WER and DR-WER were successfully improved by using the LM.
However, in the transcript-enriched model, the WER and DR-WER were not improved well with the standard LM due to the vocabulary mismatch based on the disfluency tags.
With the transcript-enriched LM, the DR-WER was recovered.

\begingroup
\renewcommand{\arraystretch}{1.1}
\begin{table}[t]
  \caption{LM integration for disfluency detection models}
  \vspace{-2mm}
  \label{tab:SWBD-LM}
  \centering
  \begin{tabular}{lcc} \hline
     & WER[\%] & DR-WER[\%] \\ \hline
    ASR only & $18.1$ & $36.9$ \\
    + standard LM & $17.0$ & $35.0$ \\ \hline
    Joint (Multi-task) & $17.8$ & $21.3$ \\
    + standard LM & $\bm{16.8}$ & $\bm{20.1}$ \\ \hline
    Joint (Transcript-enriched) & $17.9$ & $21.4$ \\
    + standard LM & $17.6$ & $21.6$ \\
    + Transcript-enriched LM & $17.3$ & $\bm{20.1}$ \\ \hline
    End-to-end removal & $-$ & $21.7$ \\
    + standard LM & $-$ & $20.6$ \\ \hline
 \end{tabular}
 \vspace{-5pt}
\end{table}
\endgroup

Finally, We confirmed that the similar trends as Switchboard were observed in CSJ, as shown in Table \ref{tab:CSJ}.
We observed that our joint models performed better than the pipeline approaches with lower latency.
Also, the multi-task model achieved comparable disfluency removed CER (DR-CER) to the transcript-enriched model while achieving lower latency.

\begingroup
\renewcommand{\arraystretch}{1.1}
\begin{table}[t]
  \caption{Streaming ASR and disfluency detection on CSJ. ``LA'' means lookahead.}
  \vspace{-2mm}
  \label{tab:CSJ}
  \centering
  \begin{tabular}{lcc} \hline
     & DR-CER[\%] & TL-Latency \\
     & eval1/2/3 & eval1 50/90 \\ \hline
    ASR only & $13.9/10.5/11.5$ & $434/880$ \\
    Joint (Multi-task) & $5.1/3.7/4.3$ & $449/876$  \\
    Joint (Transcript-enriched) & $5.1/3.8/4.2$ & $465/886$ \\
    End-to-end removal & $5.2/3.8/4.2$ & $438/810$ \\ \hline
    ASR + BERT (No LA) & $6.6/5.1/5.9$ & $486/975$  \\
    ASR + BERT (LA=2) & $5.5/4.0/4.7$ & $521/990$ \\ \hline
 \end{tabular}
 \vspace{-12pt}
\end{table}
\endgroup

\vspace{-10pt}
\section{Conclusions}
\vspace{-7pt}

In this study, we have proposed streaming joint ASR and disfluency detection.
Compared to existing pipeline approaches, joint modeling has the advantages of using acoustic information, lower latency, and a smaller memory footprint.
The transcript-enriched joint model adds special symbols to the vocabulary to indicate the disfluent part, which causes latency degradation and a failure of standard LM integration.
We propose a multi-task model that performs both token and disfluency predictions with two output layers.
To model the token-disfluency dependency, a token-dependency mechanism is employed to bridge the two output layers.
We experimentally confirmed that our joint models outperformed pipeline methods and that the multi-task model achieved competitive accuracy to the transcript-enriched model with lower latency.
In a future study, we will investigate leveraging data without disfluency annotations.

\bibliographystyle{IEEEbib}
\bibliography{strings,refs}

\end{document}